\documentclass{article}
\usepackage{ijcai23}

\usepackage{times}
\usepackage{soul}
\usepackage{url}
\usepackage[hidelinks]{hyperref}
\usepackage[utf8]{inputenc}
\usepackage[small]{caption}
\usepackage{graphicx}
\usepackage{subfig}
\usepackage{float}
\usepackage{amsmath}
\usepackage{amsthm}
\usepackage{booktabs}
\usepackage{algorithm}
\usepackage{algorithmic}
\usepackage[switch]{lineno}
\usepackage{multirow}

\usepackage[american]{babel}
\addto\extrasamerican{% only works with \usepackage[american]{babel}
}

\newlength\myindent
\newcommand\bindent{%
  \begingroup
  \setlength{\itemindent}{\myindent}
  \addtolength{\algorithmicindent}{\myindent}
}
\newcommand\eindent{\endgroup}

\title{Fin-Fed-OD: Federated Outlier Detection on Financial Tabular Data}
\author{
    Author Name
    \affiliations
    Affiliation
    \emails
    email@example.com
}

\author{Dayananda Herurkar$^{1,2}$, Sebastian Palacio$^{1}$, Ahmed Anwar$^{1}$, Jörn Hees$^{1,3}$, and Andreas Dengel$^{1,2}$ \\% <-this % stops a space 
$^{1}$German Research Center for Artificial Intelligence (DFKI), Kaiserslautern, Germany\\%
$^{2}$RPTU Kaiserslautern-Landau, Germany\\%
$^{3}$Bonn-Rhein-Sieg University of Applied Sciences, St. Augustin, Germany\\%
email: \texttt{firstname.lastname@dfki.de}
}
\usepackage[font=small]{caption}

\begin{document}

\maketitle

\begin{abstract}

Anomaly detection in real-world scenarios poses challenges due to dynamic and often unknown anomaly distributions, requiring robust methods that operate under an open-world assumption.
This challenge is exacerbated in practical settings, where models are employed by private organizations, precluding data sharing due to privacy and competitive concerns. 
Despite potential benefits, the sharing of anomaly information across organizations is restricted.
This paper addresses the question of enhancing outlier detection within individual organizations without compromising data confidentiality.
We propose a novel method leveraging representation learning and federated learning techniques to improve the detection of unknown anomalies. 
Specifically, our approach utilizes latent representations obtained from client-owned autoencoders to refine the decision boundary of inliers. 
Notably, only model parameters are shared between organizations, preserving data privacy.
% The efficacy of our proposed method is evaluated on two established anomaly detection datasets in a distributed setting.
The efficacy of our proposed method is evaluated on two standard financial tabular datasets and an image dataset for anomaly detection in a distributed setting.
The results demonstrate a strong improvement in the classification of unknown outliers during the inference phase for each organization's model.

\end{abstract}

\section{Introduction}

\begin{figure}[htp]
% \begin{figure*}[ht]
\centering
\includegraphics[width=\linewidth]{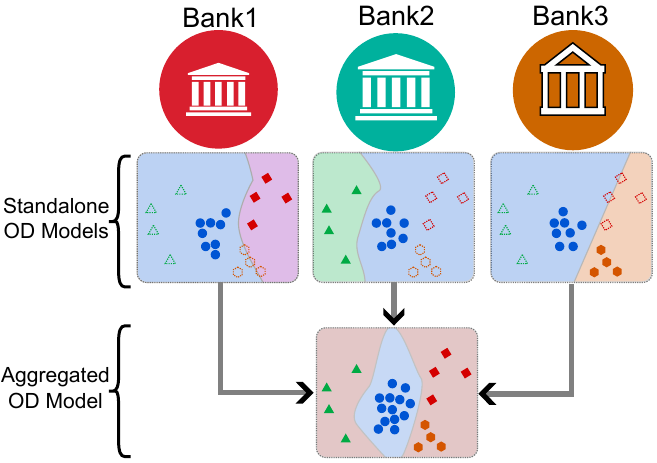}
\caption{Standalone models trained using local data (Inliers: Blue) favor detecting only known outliers (Bank1:Red, Bank2:Green, Bank3:Grey) but are not robust against unknown outliers from other organizations. The aggregated model can adapt to distinct representations and adjust its boundaries to detect unknown outliers.}
\label{fig:Intro_Prob}
\end{figure}
% \end{figure*}

Modern industries, including finance, insurance, retail, and healthcare, grapple with an escalating incidence of fraudulent and illicit transactions. 
As a result, finding financial fraud has gathered more attention recently. 
In the United States alone, the year 2019 witnessed nearly 500,000 fraud-related complaints, resulting in a staggering total loss of 3.5 billion dollars—a 30\% increase from the preceding year \cite{Fin-Fraud1}. 
% The indispensable role of outlier detection (OD) in combating fraud has been well-established.
Outlier detection (OD) has been widely used and proved highly effective for fraud detection \cite{Fin-Fraud2}.
It has been one of the most active research areas for decades, developing many OD algorithms for real-world use cases \cite{OZBAYOGLU2020106384}. 
In particular, over the past decade, the advent of Deep Learning (DL) models has introduced representation learning into OD, demonstrating promising outcomes \cite{arora2017provable}. 
Specifically, the utilization of autoencoders (AE) for representation learning has proven effective in acquiring rich representations for downstream OD tasks.

Despite these advancements, adversaries continually evolve, exploiting modern technologies to circumvent sophisticated solutions. 
This dynamic landscape poses challenges for financial institutions in apprehending perpetrators engaged in illicit activities. 
In a non-distributed setup, standalone or local models trained on an organization's specific data become vulnerable to dynamic and unknown distributions of anomalies. % the accelerating pace and sophistication of fraudulent activities.
As illustrated in \autoref{fig:Intro_Prob}, standalone OD models struggle to % adapt to unknown distributions of anomalies and cannot 
recognize or differentiate anomalies from other organizations, leading to potential false negatives or misclassifications.
Furthermore, locally trained models exhibit inherent biases.  
Therefore, to sustain the open-world classification problem, there is a pressing need to enhance the robustness of OD models to effectively address known and emerging outliers.

Addressing these challenges leverages collaboration among multiple organizations to amalgamate their knowledge of various fraudulent cases within their datasets. 
However, privacy and confidentiality concerns present huge obstacles to such collaboration. 
Federated Learning (FL)~\cite{fl-mcmahan} emerges as a viable solution, introducing a new distributed machine learning (ML) paradigm where multiple clients (organizations) collaboratively train an ML model. %without sharing proprietary data.
In FL, individual client data samples are not shared with other participating clients; instead, local model updates are transmitted to a centralized server for aggregation. 
The resultant aggregated (global) model, embodying collective knowledge, is then disseminated back to each participating client. 
The fundamental principle of the FL approach lies in sharing local model updates while safeguarding sensitive data.

In this study, we integrate FL with representation learning to devise a robust OD approach capable of handling both known and unknown outliers. 
Our methodology involves using AE as a client-local model, trained on local data and aggregated at the server to form a global model. 
The aggregated global model is subsequently returned to participating clients, replacing their local models. 
We extract latent space representations from trained AE models for all data samples, which are then fed into an OD model for outlier detection. 
% Our approach excels in detecting unknown outliers compared to standalone models (non-distributed) and maintains its performance in detecting known outliers without compromising efficiency.
This setup facilitates the convergence of distinct representations, priming local models to adapt and tighten the boundaries of their inlier distribution, thereby making them more robust to unknown outliers.

% In summary, we present the following contributions:
% \begin{itemize}
%     \item We synergize representation learning and FL, exploring the untapped potential of this combination to develop a robust outlier detection approach.
%     \item We showcase the effectiveness of representation learning in OD models, emphasizing its capability to detect unknown outliers over standalone models. Our approach strikes a balance between generalization and personalization, enabling the model not only to distinguish outliers from inliers but also to differentiate outliers originating from different clients.
%     \item Through extensive experimentation, we demonstrate the algorithm-agnostic nature of our approach.
% \end{itemize}

In summary, our primary contribution involves \textit{the integration of representation learning and Federated Learning (FL), yielding a robust outlier detection approach called Fin-Fed-OD}. 
This synergistic combination demonstrates the effectiveness of representation learning in OD models, enhancing the detection of unknown outliers compared to standalone models and achieving a balance between generalization and personalization. 
As a result, our proposed approach not only exhibits the ability to distinguish outliers from inliers but also differentiates outliers originating from various clients. 
Furthermore, our extensive experimentation establishes the algorithm-agnostic nature of this approach, showcasing its adaptability across different OD algorithms.

The remainder of this work is structured as follows: \autoref{sec: related_work} provides a review of relevant literature, highlighting gaps in the current state-of-the-art. 
In \autoref{sec: approach}, we detail our approach to federated OD using representation learning. 
The experimental setup, model architecture, datasets used, and evaluation measures are described in \autoref{sec:exp_setup}, while \autoref{sec: res} presents the results and comparisons. 
We conclude in \autoref{sec: conclusion}, summarizing our main findings and identifying opportunities for future research.

\section{Related Work} \label{sec: related_work}

% \begin{markdown}
%     - Outlier Detection in FL
%     - Representation Learning using AE for OD
%     - How is our approach different from the above?
% \end{markdown}

% \subsection{Outlier Detection in FL}
% \subsection{Representation Learning using AE for OD}

The persistent challenges posed by fraudulent transactions in industries such as finance have stimulated substantial research efforts \cite{AI-in-Finance}, \cite{Fin-Fraud2}, \cite{RECol}, \cite{Fin-Fraud3}. 
Research in this area spans diverse financial tasks, including insurance, retail, risk assessment, and credit \cite{OZBAYOGLU2020106384}. 
This section provides a comprehensive survey of the existing literature, encompassing outlier detection (OD) techniques for financial tabular data and OD within a federated learning (FL) framework.

\noindent \textbf{Outlier Detection in Financial Tabular Data:}
Outlier detection has been a focal point of research across various domains for decades, with an intensified focus in the financial domain \cite{Fin-Fraud1}. 
Given the prevalence of tabular data in financial tasks, especially with the evolution of Deep Learning (DL), this domain has become a promising application area for recent methods \cite{Borisov-2022}. 
Autoencoders (AE), in particular, have gained widespread usage for outlier detection in financial data \cite{chalapathy2019deep}. 
Numerous techniques leveraging autoencoders have been developed, ranging from detecting anomalies in accounting data \cite{Marco2} to explaining anomalies in financial tabular data \cite{Timur}. AE applications extend to transmuting anomalies between tabular data formats \cite{Herurkar-ijcnn23}, learning behavioral fraud \cite{wedge2017solving}, and identifying anti-money laundering \cite{Zahra}. %,\cite{Ebberth}. 
Autoencoders also prove effective in detecting credit card fraud \cite{Pumsirirat2018}, \cite{Zahra}. % \newline

\noindent \textbf{Representation Learning:}
Representation learning is a process of extracting features or learning abstract representations of input data to solve a particular task \cite{Bengio}.
Such representation features contain valuable and quality information, which makes it easier to perform tasks such as classification or prediction \cite{arora2017provable}.
There are many techniques for representation learning used for financial tasks involving, for example, clustering analysis \cite{Thiprungsri2011ClusterAF}, business process mining \cite{business-process}, transaction mining \cite{transaction}, naive bayes classification \cite{nbclassifier}, and network analysis \cite{network-analysis}.
Nowadays, there are novel representation learning methods based on DL models; the popular technique is autoencoders.
By taking advantage of autoencoders' data compression technique into lower dimension feature space called latent space, many approaches could show state-of-the-art performance for outlier detection tasks.
\cite{memae} showed limitations of autoencoders that generalize well for outlier samples, leading to bad OD performance.
Hence, they proposed a memory-augmented autoencoder called MemAE, which updates the contents in latent space to get an improved outlier detection output.
Similarly, \cite{math10010112} uses the unlabelled samples from different domains as negative samples and then shapes the latent layer to boost autoencoder performance in classifying the outliers. 
\cite{latentout} % and \cite{sparce} 
utilize the advantage of combined latent and reconstruction error space and show that outliers tend to lie in the sparsest regions, which helps to detect complex outliers.
\cite{DAGMM} also explores a similar direction to show the benefits of the latent space representation in OD.
Applied representation learning has also gained success in recent times, encompassing variational autoencoders \cite{latentout}, % \cite{sparce}, 
adversarial autoencoders % \cite{robust}, 
\cite{Marco1}, and latent space generation approaches \cite{attention}. % \newline

% \begin{figure}[htp]
\begin{figure*}[ht]
\centering
\includegraphics[width=\linewidth]{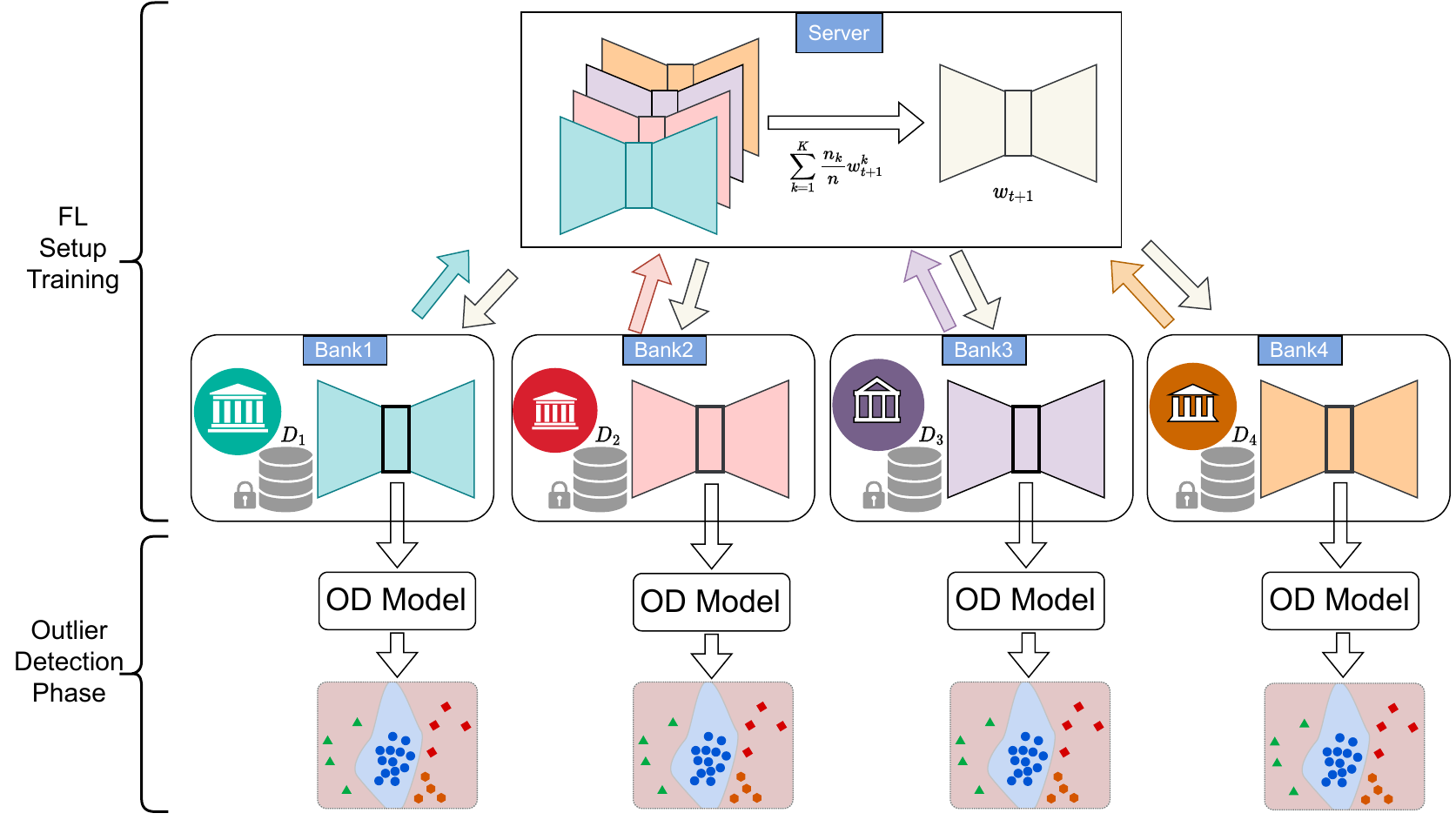}
\caption{FL-OD Setup}
\label{fig:FL_OD_Setup}
% \end{figure}
\end{figure*}

\noindent \textbf{Outlier Detection using Federated Learning:}
Federated Learning (FL), introduced by \cite{fl-mcmahan}, empowers distributed clients to train local models using their respective data, aggregating these models centrally under the coordination of a central server. 
This collaborative approach allows clients to learn a common model without sharing their private data \cite{FederatedOD}. 
FL has found applications in outlier detection and various tasks across domains. 
For instance, \cite{smart-buildings} combined FL with outlier detection, employing an LSTM-based model to detect anomalies in IoT sensor data for smart buildings. 
Communication-efficient on-device FL approaches for anomaly detection in IoT time series data have been proposed by \cite{communication}, and \cite{enhancing} introduced an FL-based approach to enhance anomaly detection in IoT data. 
FL, however, faces challenges from malicious attacks, prompting studies that integrate FL with outlier detection to mitigate such effects \cite{abnormal}, \cite{malicious}. 
Additionally, \cite{byzantine} utilized an anomaly detection approach to counter Byzantine attacks, proposing a defense mechanism, while \cite{attack-adaptive} presented a robust FL approach with attack-adaptive aggregation. 
Numerous approaches combining autoencoder-based outlier detection with FL have been proposed for intrusion detection \cite{intrusion}, conditional variational autoencoders \cite{cvae}, fast deep autoencoders \cite{fast}, vanilla autoencoders \cite{vanilla-ae-fl}, and for attacks+defenses in FL \cite{free-riders}. % \newline

While existing literature emphasizes the popularity of autoencoder-based representation learning for outlier detection and various FL-based outlier detection approaches, few studies have explored the application of representation learning with FL, particularly for financial tabular data. 
This study addresses this gap by demonstrating the advantages of representation learning in FL and showcasing how OD models, trained using this approach, exhibit robustness in detecting new and sophisticated outliers.

\section{Approach} \label{sec: approach}

% \begin{markdown}
%     - FL Model Architecture + Fig
%     - How are outliers detected? (How are ML algorithms trained?)
% \end{markdown}

This section will first elaborate on the standalone (baseline) models for outlier detection. Then, we proceed to expand those models for the federated learning approach.

\subsection{Standalone (Baseline) Model For Outlier Detection} \label{subsec:local_model}
We opt for the popular representation learning based approach for outlier detection and utilize autoencoder neural networks (AE) for this purpose \cite{memae}, \cite{latentout}, \cite{Marco1}.
% Broadly, AE is made up of two nonlinear functions known as encoder and decoder networks.
An AE network represented as $d(e(X))$ comprises two parts: encoder $f_\theta$ and decoder $g_\psi$. 
Given an input data `X', the encoder performs data compression into a lower dimensional feature space (latent space), typically denoted as \(h\), and then the decoder accomplishes the reconstruction, i.e, reconstructs \(h\), back to the original data space.
The complete network is trained in an end-to-end unsupervised fashion by minimizing the reconstruction loss ($L$), formally defined as follows:

\begin{equation}
	\arg\min_{\theta, \psi} \|X - g_\psi(f_\theta(X))\|
	\label{equ:ae_objective}
\end{equation}

Upon successful training of AE, the latent space representation of data $h = f_{\theta}(X)$ is extracted and then passed as input to a standard OD algorithm for detecting outliers.
These latent representations (\(h\)) contain rich and valuable information that is beneficial for many downstream tasks and are often used for tasks like outlier detection and clustering approaches \cite{arora2017provable}.

% \subsection{Federated Learning For Outlier Detection} \label{subsec:fl_model}
\subsection{Collaborative Federated Learning Environment (CoFLE)} \label{subsec:fl_model}
To enhance the robustness of our OD model against unknown outliers from different clients while preserving the confidentiality of each client's data, we extend the baseline models to a Federated Learning (FL) setup \cite{fl-mcmahan}. 
In this section, we provide a detailed description of our FL-based outlier detection approach.
Our FL setup consists of K individual clients (e.g., corresponding to K different banks). 
Each client possesses its proprietary dataset represented by ${D}_i$, where $i\in{\{1,...,K\}}$, containing different types of outlier samples.
Additionally, each client has its own AE model represented as ${w}_i$ trained on its local dataset ${D}_i$.
In real-world scenarios, datasets $D_i$ are private and only accessible to the respective client. 
% As a result, $D_i$ and $D_j$ can be non-independent and identically distributed (non-iid).
% FL setup necessitates the use of a central server and a certain number of clients. 
The implementation of FL setup begins with the utilization of a central server in conjunction with a certain number of clients.
At first, the server initializes its global model $w_0$ and randomly selects C, a subset of the K available clients. 
To initialize the FL, the sever broadcasts its global model to selected C clients.
At each communication round r out of a maximum of R rounds, each selected client conducts the local training of their model ($w_i$) using their private dataset $D_i$.
After successful training, each client sends their model parameters to the server.
Finally, the server aggregates all the received updates 
% \begin{equation}
%     AE_{CoFLE} = \psi(w_1, w_2, w_3, w_4) where \psi(\cdot) is an aggregation function
%     \vspace{2mm}
%     \label{equ:fl_aggregation}
% \end{equation}
and replaces the global model with the updated parameters: $AE_{CoFLE} = \psi(w_1, w_2, w_3, w_4)$, where $\psi(\cdot)$ is an aggregation function.
This process repeats until the final communication round R is reached.
Our FL approach is explained in Algorithm\autoref{alg:FL_algo}, and \autoref{fig:FL_OD_Setup} shows the setup.
Latent representation from each client model is extracted and fed to the client's OD model to detect outliers like in \autoref{subsec:local_model}.

\section{Experimental Setup} \label{sec:exp_setup}

% \begin{markdown}
%     - Datasets Description + Table
%     - Synthetic Outliers and Real Outliers setup
%     - Baseline models
%     - FL+AE+ML Hyperparameters + Table
%     - Evaluation Metrics
% \end{markdown}

In this section, we explain the details of our experimental setup and datasets used for training.
We describe the different noise injection methods applied to create different types of outliers on tabular datasets and the procedure to create multiple image datasets with natural outliers.
The PyTorch v$1$ \cite{pytorch} framework was used to train and evaluate the autoencoder neural network models.

\subsection{Tabular Dataset + Synthetic Outliers} \label{subsec:tab_data}
We employed two standard financial tabular datasets for outlier detection to evaluate our approach.
Both the tabular datasets have mixed-type (numerical + categorical) attributes. % and are listed in Table 1.
As a part of data pre-processing, all the categorical features are one-hot encoded, and numerical features are standardized (mean=0 and std=1).
The description of each dataset is given below:

\begin{itemize}
    \item \textbf{Credit Default}\footnote{Publicly available via: \url{https://archive.ics.uci.edu/ml/datasets/default+of+credit+card+clients}} \cite{UCI}: This dataset contains information on bill payment of credit card clients of Taiwan from April to September 2005. It also has more information, such as payment history, demographic factors of clients, and their default payments.
    
    \item \textbf{Adult}\footnote{Publicly available via \url{https://archive.ics.uci.edu/ml/datasets/adult}}: The dataset is taken from the UCI repository \cite{UCI}. It consists of individuals' income records and personal information like education, marital status, etc. A person with an income exceeding \$$50$k per year is labeled an outlier. 
\end{itemize}

\begin{algorithm}
\caption{FL-OD Setup. The K clients are indexed by k, D is the local datasets indexed by i, E is the number of local epochs, B denotes the local batch size, C $\in$ (0, 1] denotes a fraction, and $\nu$ is the learning rate} \label{alg:FL_algo}
\begin{algorithmic}
\STATE Split data $D_{i}$ into $D_{i}^{train}$ and $D_{i}^{test}    \forall i\in{\{1,..,K\}}$
\STATE \textbf{FL Setup:}
\bindent
\STATE \textbf{Server Process:}
\bindent
\STATE Initialize $w_0$.
\FOR{each round r = 1,2,3.. R }
\STATE $ m \gets $ max($C.K$, 1)
\STATE $S_t \gets $ random set of m clients
\FOR{each client $k \in S_t$ \textbf{in parallel}}
\STATE ${w}_{t+1}^{k} \gets $ ClientUpdate($k, w_t$)
\ENDFOR
\STATE ${w}_{t+1} \gets \sum_{k=1}^{K} \frac{n_k}{n} {w}_{t+1}^{k}$
\ENDFOR
\eindent

\STATE
\STATE \textbf{ClientUpdate(k, w):}
\bindent
\STATE $B \gets $ split local train data into batches of size B
\FOR{each local epoch i from 1 to E}
\FOR{batch $b \in $B}
\STATE $w \gets w - \nu \Delta$(w;b)
\ENDFOR
\ENDFOR
\STATE return $w$ to the server
\eindent
\eindent
\STATE

\STATE \textbf{Outlier Detection Process:}
\bindent
\STATE Once FL setup training is complete
\FOR{each client $k \in K$}
\STATE Extract Latent representation of $D_{k}^{train}$ as $Repr_{k}^{train}$ 
\STATE $Repr_{k}^{train} = AE_{k}(D_{k}^{train})$
\STATE $ODModel_{k} \gets train(Repr_{k}^{train})$
\STATE $Output_{k} \gets ODModel_{k}(Repr_{i}^{test}) \forall i\in{\{1,..,K\}}$
\ENDFOR
\eindent

\end{algorithmic}
\end{algorithm}

\noindent \textbf{Synthetic Outlier Generation:} To evaluate the model's robustness against known and unknown outliers, we created specific synthetic outliers for each client dataset. 
As publicly available datasets labeled with different types of outliers are scarce, we adopted standard practices of corrupting clean data to generate artificial outliers \cite{Timur}, \cite{cell_corrupt1}, \cite{cell_corrupt2}.
Four client datasets were created from each tabular dataset by picking only clean data, with a substantial proportion of each dataset being dissimilar.  
We select 3\% of inliers from each created dataset at random and convert them into outliers by corrupting randomly selected attribute (column) values.
We applied four artificial corruption techniques to generate four types of outliers (two for categorical and two for numerical features): one specific for each client dataset.

\noindent \textbf{Numerical Features: } In each sample selected for corruption, 25\% of numerical features are selected uniformly at random.
On each selected feature, noise $\delta$ is added to turn the clean sample into an outlier such as $\tilde{x}_{n}^{d} = x_{n}^d + \delta $ where $n$ is a sample index, and $d$ represents a feature or a column.
$\delta$ is either randomly sampled from Gaussian, Laplace, or Log-Normal distributions with $\mu=0$ and $\sigma=\sigma_d \gamma $ or by adding random high values to imitate real-world outliers (ex: mismatch currency). 
Selection of $\gamma$ follows uniform distribution $\gamma=\text{Unif}(3,5)$ and $\sigma_d$ is the standard deviation of the original attribute.

\noindent \textbf{Categorical Features: } We follow a similar strategy where 25\% of categorial features are selected uniformly at random for corruption, and two methods are used.
In the first one, we replace the original value with a new categorical value created by manipulation. 
In the second one, the original value is replaced by a least frequently occurring categorical value of the selected attribute.  

\subsection{Image Dataset + Natural Outliers} \label{subsec:image_data}
We benchmarked our approach on detecting real outliers using a standard MNIST dataset \cite{MNIST}.
To turn MNIST into a dataset for outlier detection, we employed the below procedure:
We chose all images with digit-0 as inliers and created four client datasets out of it \cite{deep-one-class}.
The majority of each client dataset contains different digit-0 images, and only a minor part of the dataset contains the same samples between clients.
To make the problem more challenging for our approach, we analyzed the classification result for MNIST dataset and chose those digits as outliers most commonly misclassified as digit-0.
So, we chose digits 3,4,6,8 as outlier samples and added one of the digits into each client dataset as an outlier.
Only 3\% of samples in each dataset are outliers.

\subsection{Model Parameters}
For all our experiments, we used AE neural network with symmetrical encoder $f_\theta$ and decoder $g_\psi$ architecture.
% Table 2 describes the model architecture and other parameters used for each dataset for training.
For both Credit Default and Adult Data, we use one hidden layer of size 128, and the latent space size is 64, whereas for MNIST (we replicated the parameters from~\cite{DASVDD}), one hidden layer of size 1024 and the latent space size is 256.
For both the tabular datasets, we apply Leaky-ReLU non-linear activations with scaling factor $\alpha$ = 0.4 and Tanh activations for MNIST dataset.
For optimizing the model parameters, we used the Adam optimizer \cite{adam-optimizer} with $\beta_{1}=0.9$, $\beta_{2}=0.999$ and optimized the AE based on the loss function in \autoref{equ:ae_objective}, which is expanded per sample as below:

\begin{equation}
    \mathcal{L}(x;\hat{x}) =   \sum_{d \in D_\text{cat}} \mathcal{L}^\text{BCE}(x^d;\hat{x}^d) +  \sum_{d \in D_\text{num}}
    \mathcal{L}^\text{MSE}(x^d;\hat{x}^d),
    \vspace{2mm}
    \label{equ:reconstruction_loss_details}
\end{equation}

\noindent where ${x}$ represents the original sample, $\hat{x}$ denotes the reconstructed sample, and $d$ represents the attribute.
Two tabular datasets with a mixed-type nature of data, we define the reconstruction loss of every instance as the sum of two losses.
For each one-hot encoded representation of the categorical attribute, the (1) binary cross entropy loss ($\mathcal{L}^\text{BCE}$) is calculated, and (2) the mean-squared loss ($\mathcal{L}^\text{MSE}$) used for the numerical attributes.
However, for MNIST, the loss value consists of only mean-squared error ($\mathcal{L}^\text{MSE}$).
Our FL setup comprised four clients and was trained for communication rounds (R) equal to 100, 200, and 100 for the Credit Default, Adult Data, and MNIST datasets, respectively. 
% FL optimization task is formally shown as:
% \begin{equation}
% 	\min L(X) = \frac{1}{C} \sum_{i=1}^{C} L_{i}(X)
% 	\label{equ:fl_obejctive}
% \end{equation}
% where $L_{i}$ is the loss function for client `i' represented as \autoref{equ:reconstruction_loss_details}.
Two aggregation methods ($\psi(\cdot)$) for client model updates were applied: Federated Averaging (FedAvg) \cite{fl-mcmahan} and FedProx FedProx \cite{fed-prox}.
Each client's AE model underwent local training for 500, 500, and 300 epochs for the three datasets, respectively.
Upon completion of AE model training, the latent representation ($h$) for dataset samples from each client was extracted and subsequently employed as input for classification algorithms aimed at detecting outliers. 
Two variations of the model were assessed for OD purposes, namely Random Forest and Multi-Layer Perceptron (MLP). 
Additionally, the effectiveness of our approach was evaluated by replacing AE by alternative deep anomaly detection methods, such as DAGMM \cite{DAGMM} and MemAE \cite{memae}.

\subsection{Evaluation Metric}
To quantitatively measure the outlier detection performance of fine-tuned models we used average precision score.
It is formally defined as 

\begin{equation}
    \text{AP} = \sum_{i=1}^{N} (R_i - R_{i-1})P_i
    \label{eq:map}
\end{equation}

\noindent where $R_i = TP / (TP + FN)$  denotes the detection recall and $P_i = TP / (TP + FN)$ denotes the detection precision of the $i-th$ anomaly score threshold.

\section{Results} \label{sec: res}
% \begin{markdown}
%     - Output Comparison table or bar plots + explain
%     - Rec error comparison table + explain
% \end{markdown}

This section presents the outcomes of our experiments, offering insights into the effectiveness of our approach through both quantitative and qualitative evaluations. 
The robustness of our method against diverse outlier types is also highlighted.

\subsection{Quantitative Evaluation}
\begin{figure}[htp]
% \begin{figure*}[ht]
\centering
\includegraphics[width=\linewidth]{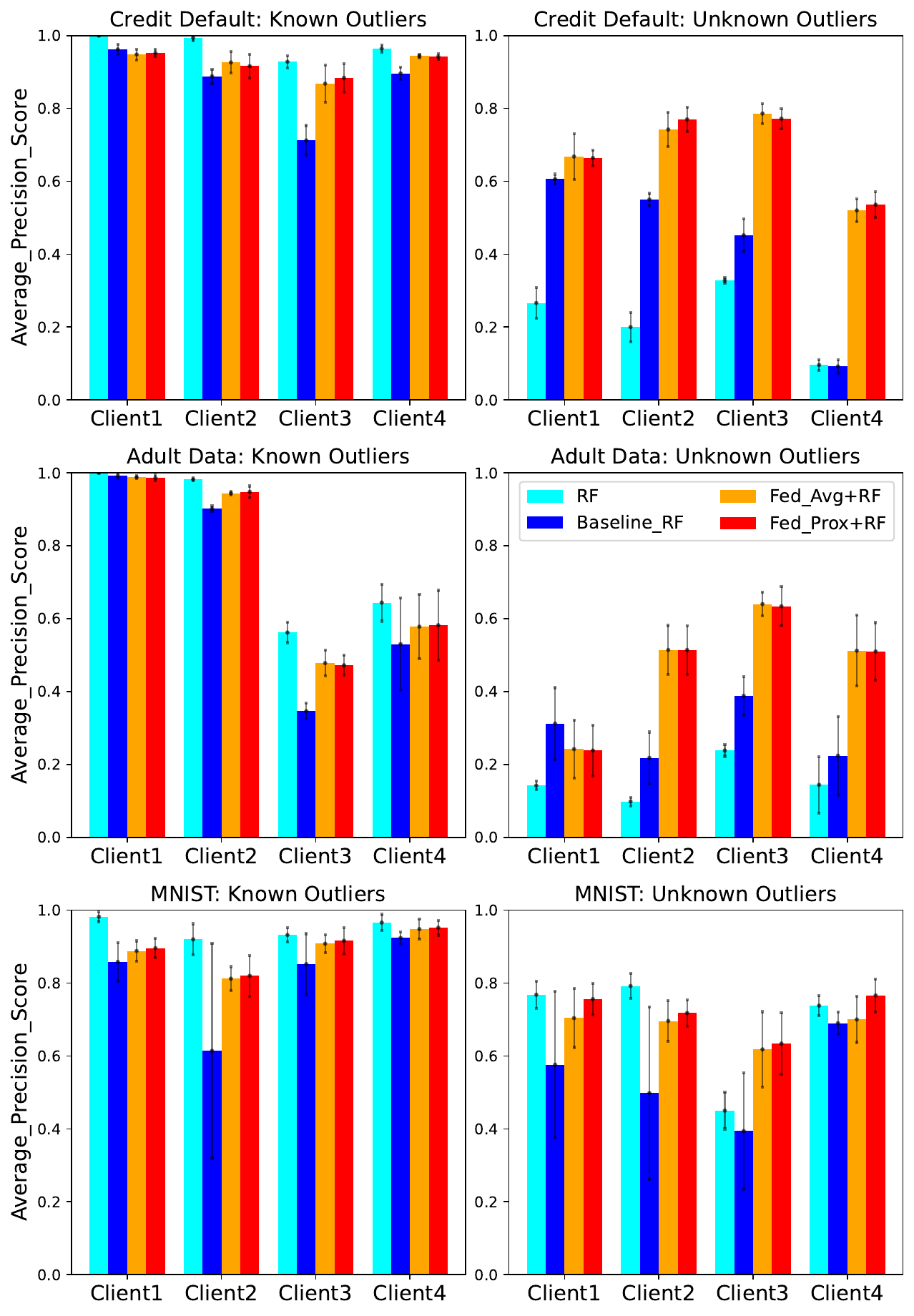}
\caption{Results of OD models clientwise. FL-OD models (Fed\_Avg+RF, Fed\_Prox+RF) outperform the baseline (Baseline\_RF) in detecting unknown outliers. %for all clients. 
Every score reflects the mean and standard deviation from five experiments.}
\label{fig:Res_Plots}
\end{figure}

In this section, we will gather multiple research questions and try to find the answers with our approach.   
We evaluated the FL setup by comparing the capabilities of OD models using FL [\autoref{subsec:fl_model}] to baseline OD models (OD models without FL setup) [\autoref{subsec:local_model}].
\autoref{fig:Res_Plots} shows the comparison of clientwise average precision score obtained for OD models across three datasets explained in \autoref{subsec:tab_data} and \autoref{subsec:image_data}.
\autoref{tab:res_table} shows the same results but average output across all clients.
We used two OD algorithms, namely Random Forest and MLP. 
Both algorithms' results are shown separately in \autoref{tab:res_table}.

\noindent \textit{``RQ1: Does the integration of representation learning enhance the performance of OD models in identifying unknown outliers?"}.

\noindent When we compare the results of OD models in detecting unknown outliers without and with representation learning, i.e., between `RF' and `baseline\_RF' % and between `MLP' and `baseline\_MLP' 
in both \autoref{tab:res_table} and \autoref{fig:Res_Plots}, the latter performs better than the first one.
`RF' is a random forest model trained using original datasets. 
In `baseline RF', we train the AE using original datasets and then extract the latent space representation (\(h\)) of AE and finally, a random forest is trained using \(h\), i.e., baseline models as discussed in \autoref{subsec:local_model}.
% The output says that there is an improvement in detecting unknown outliers by OD models when incorporated with representation learning.
The findings demonstrate a notable improvement in the detection of unknown outliers when representation learning is integrated into OD models.

% \end{figure*}

\noindent \textit{``RQ2: Will the combination of FL and representation learning benefit OD models in detecting unknown outliers?"}.

\noindent We compared the result between OD models using FL setup and baseline, i.e., `Fed\_Avg+RF', `Fed\_Prox+RF', (explained in \autoref{subsec:fl_model}) and `baseline\_RF' (explained in \autoref{subsec:local_model}).
We tested our approach for two standard FL algorithms, `Fed\_Avg' and `Fed\_Prox'.
In \autoref{tab:res_table}, we can see that both `Fed\_Avg+RF' and `Fed\_Prox+RF' outperformed `baseline\_RF' in detecting unknown outliers.
Similar results can be seen in \autoref{fig:Res_Plots} that even by comparing results at a client level, the FL models outperformed baseline OD models.
Another notable result is that training OD models in FL setup also improves the result of detecting known outliers.
So, combining information from other clients has a positive impact on detecting both known and unknown outliers.
We clarify these results; we also ran experiments by replacing RF with MLP (Multi Layer Perceptron) as an OD model.
We obtained similar results where FL based MLP models `Fed\_Avg+MLP' and `Fed\_Prox+MLP' outperformed `baseline\_MLP' in detecting unknown outliers but without decreasing its performance on known outliers.
To test the robustness of our approach, we replaced the AE with alternative OD models, specifically, DAGMM and MemAE. 
The repeated experiments, detailed in \autoref{tab:res_table}, demonstrate that our proposed approach consistently enhances the performance of detecting both known and unknown outliers in DAGMM and MemAE models.
The above thorough analysis answers our RQ2 that ``FL-OD models are robust against detecting unknown outliers". 
To bolster these findings, we conducted experiments with alternative evaluation metrics such as 'F1-score,' 'PR-AUC,' and 'ROC-AUC,' yielding consistent outcomes, and results from alternative metrics can be found in the appendix.

\begin{table}
\begin{tabular}{@{}llrr@{}}
\toprule
\multicolumn{1}{c}{\textbf{Data}} & \multicolumn{1}{c}{\textbf{Model}} & \multicolumn{1}{c}{\textbf{\begin{tabular}[c]{@{}c@{}}Known\\ Outliers\end{tabular}}} & \multicolumn{1}{c}{\textbf{\begin{tabular}[c]{@{}c@{}}Unknown\\ Outliers\end{tabular}}} \\ \midrule
\multirow{8}{*}{\begin{tabular}[c]{@{}l@{}}Credit \\ Default\end{tabular}} & RF & \textbf{0.971 $\pm$ 0.005} & 0.222 $\pm$ 0.005 \\
 & Baseline\_RF & 0.864 $\pm$ 0.007 & 0.425 $\pm$ 0.010 \\
 & Fed\_Avg+RF & 0.921 $\pm$ 0.013 & \textbf{0.679 $\pm$ 0.026} \\
 & Fed\_Prox+RF & 0.923 $\pm$ 0.015 & \textbf{0.685 $\pm$ 0.012} \\ \cmidrule(l){2-4} 
 & MLP & \textbf{0.953 $\pm$ 0.006} & 0.240 $\pm$ 0.013 \\
 & Baseline\_MLP & 0.941 $\pm$ 0.004 & 0.197 $\pm$ 0.020 \\
 & Fed\_Avg+MLP & 0.941 $\pm$ 0.013 & \textbf{0.369 $\pm$ 0.050} \\
 & Fed\_Prox+MLP & 0.933 $\pm$ 0.007 & \textbf{0.375 $\pm$ 0.014} \\ \cmidrule(l){2-4} 
 & DAGMM & 0.483 $\pm$ 0.017 & 0.365 $\pm$ 0.018 \\
 & DAGMM+Ours & \textbf{0.505 $\pm$ 0.025} & \textbf{0.368 $\pm$ 0.013} \\ \cmidrule(l){2-4} 
 & MemAE & 0.560 $\pm$ 0.011 & 0.361 $\pm$ 0.010 \\
 & MemAE+Ours & \textbf{0.609 $\pm$ 0.024} & \textbf{0.583 $\pm$ 0.015} \\ \midrule
\multirow{8}{*}{\begin{tabular}[c]{@{}l@{}}Adult \\ Data\end{tabular}} & RF & \textbf{0.797 $\pm$ 0.012} & 0.155 $\pm$ 0.012 \\
 & Baseline\_RF & 0.692 $\pm$ 0.030 & 0.285 $\pm$ 0.025 \\
 & Fed\_Avg+RF & 0.747 $\pm$ 0.023 & \textbf{0.477 $\pm$ 0.018} \\
 & Fed\_Prox+RF & 0.747 $\pm$ 0.029 & \textbf{0.474 $\pm$ 0.022} \\ \cmidrule(l){2-4} 
 & MLP & \textbf{0.762 $\pm$ 0.024} & 0.219 $\pm$ 0.021 \\
 & Baseline\_MLP & 0.745 $\pm$ 0.029 & 0.153 $\pm$ 0.012 \\
 & Fed\_Avg+MLP & 0.754 $\pm$ 0.020 & \textbf{0.376 $\pm$ 0.022} \\
 & Fed\_Prox+MLP & 0.752 $\pm$ 0.022 & \textbf{0.381 $\pm$ 0.036} \\ \cmidrule(l){2-4} 
 & DAGMM & 0.391 $\pm$ 0.056 & 0.230 $\pm$ 0.056 \\
 & DAGMM+Ours & \textbf{0.396 $\pm$ 0.051} & \textbf{0.234 $\pm$ 0.020} \\ \cmidrule(l){2-4} 
 & MemAE & 0.413 $\pm$ 0.016 & 0.291 $\pm$ 0.010 \\
 & MemAE+Ours & \textbf{0.686 $\pm$ 0.041} & \textbf{0.498 $\pm$ 0.023} \\ \midrule
\multirow{8}{*}{MNIST} & RF & \textbf{0.950 $\pm$ 0.010} & 0.687 $\pm$ 0.010 \\
 & Baseline\_RF & 0.812 $\pm$ 0.083 & 0.539 $\pm$ 0.080 \\
 & Fed\_Avg+RF & 0.889 $\pm$ 0.010 & 0.679 $\pm$ 0.065 \\
 & Fed\_Prox+RF & 0.932 $\pm$ 0.021 & \textbf{0.721 $\pm$} 0.032 \\ \cmidrule(l){2-4} 
 & MLP & \textbf{0.932 $\pm$ 0.005} & 0.609 $\pm$ 0.012 \\
 & Baseline\_MLP & 0.874 $\pm$ 0.114 & 0.523 $\pm$ 0.116 \\
 & Fed\_Avg+MLP & 0.912 $\pm$ 0.018 & 0.571 $\pm$ 0.087 \\
 & Fed\_Prox+MLP & 0.931 $\pm$ 0.019 & \textbf{0.621 $\pm$ 0.063} \\ \cmidrule(l){2-4} 
 & DAGMM & 0.665 $\pm$ 0.040 & 0.355 $\pm$ 0.041 \\
 & DAGMM+Ours & \textbf{0.676 $\pm$ 0.043} & \textbf{0.416 $\pm$ 0.113} \\ \cmidrule(l){2-4} 
 & MemAE & \textbf{0.979 $\pm$ 0.006} & 0.630 $\pm$ 0.035 \\
 & MemAE+Ours & 0.972 $\pm$ 0.006 & \textbf{0.736 $\pm$ 0.009} \\ \bottomrule
\end{tabular}
\caption{Average Precision Score of baseline and FL-OD Models. OD Models using FL setup outperform others in detecting unknown outliers and don't lose performance on detecting known outliers as well. % So FL-OD models are robustness against unknown outliers. 
Every score reflects the mean and standard deviation from five experiments.}
\label{tab:res_table}
\end{table}

\subsection{Qualitative Evaluation}

\begin{figure}[htp]
% \begin{figure*}[ht]
\centering
\includegraphics[height=1.1\linewidth, width=\linewidth]{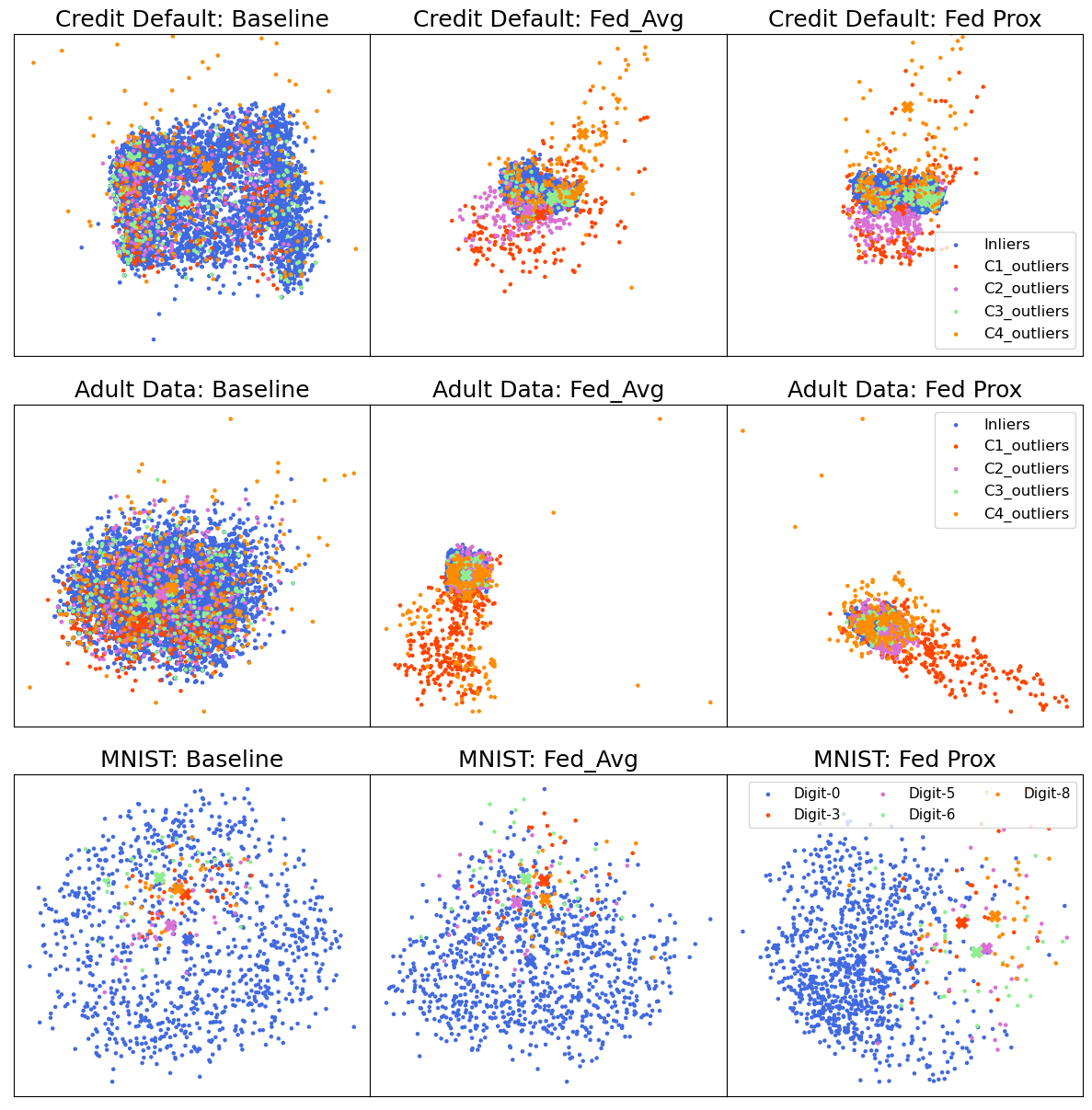}
\caption{Latent Space Visualization of Baseline, Fed\_Avg, and Fed\_Prox models}
\label{fig:Latent_Space_Vis}
\end{figure}
% \end{figure*}

To elucidate the superior performance of FL-OD models over the baseline, we examine the latent space representations of AE with and without FL setup. 
% These representations are extracted and subjected to Principal Component Analysis (PCA) \cite{pca} for visualization, as depicted in \autoref{fig:Latent_Space_Vis} for all three datasets used in our experiments.
As depicted in \autoref{fig:Latent_Space_Vis}, the latent representations for all three datasets are extracted and subjected to Principal Component Analysis (PCA) % \cite{pca} 
for visualization.
Comparison of plots for % latent space representations 
baseline and FL setup reveals that FL models exhibit superior outlier isolation by consolidating them into distinct clusters, as opposed to the sparse representation offered by baseline models. 
The compact grouping of samples by FL models facilitates easy classification of both known and unknown outliers.
% This observation suggests that classification of outliers can be achieved not only by other machine learning algorithms but also by clustering algorithms when integrated with FL setup.
% This observation suggests that the classification of outliers can be achieved by other machine learning algorithms when integrated with the FL setup.

An essential insight from \autoref{fig:Latent_Space_Vis} is that FL models can generate distinct clusters for outliers from different clients. 
This capability enables the classification of outliers into various types or the assignment of outliers to specific clients. 
Moreover, the visualization indicates that FL models strike a balance between generalization and personalization, a feature lacking in baseline models. 
Generalization, represented by maintaining more samples within the same class clusters, and personalization, facilitating the identification of client-specific outliers, are achieved more effectively by FL models.
In summary, the qualitative evaluation underscores the efficacy of FL models in creating meaningful and well-separated clusters in the latent space, providing a solid foundation for improved outlier detection.

\section{Conclusion and Future Work} \label{sec: conclusion}
% \begin{markdown}
%     - Improving Latent-FL
%     - Measure the effect of Fairness 
% \end{markdown}

In this study, we introduced a novel approach to discern outliers in mixed-type tabular data originating from diverse clients. 
To enhance the efficacy of our methodology, we integrated two key techniques: representation learning for outlier detection (OD) and federated learning (FL) for merging insights from different clients. 
Our assessment was conducted on two financial tabular datasets featuring synthetic outliers and an image dataset with real outliers. 
The experimental findings present evidence supporting the notion that our approach facilitates collaborative learning among clients without the need for direct data sharing. 
Consequently, it yields a robust OD model proficient in identifying unknown outliers more effectively than standalone models. 
Notably, our extended experiments affirm the method's versatility, showcasing its compatibility with various OD algorithms.
The qualitative evaluation offered insights into how representation learning augments the effectiveness of OD models within the FL framework. Our developed OD models exhibit a nuanced equilibrium between generalization and personalization capabilities.

Looking ahead, we envision further analysis of this framework to empower financial institutions in detecting unknown outliers and gaining insights into diverse outlier types without compromising proprietary data security. 
Subsequent investigations could delve into the exploration of latent layers, fostering the development of latent layer-based FL models for enhanced outlier detection. 
Additionally, assessing the impact of fairness within the FL context could provide valuable insights for refining and optimizing the proposed framework.

\newpage
\bibliographystyle{named}
\bibliography{ijcai23}

\onecolumn
\section*{\centering Appendix}

\section{Quatitative Evaluation Result}
% We have repeated our quantitative evaluation using different evaluation metrics such as `F1-Score', `PR-AUC', and `ROC-AUC'.
% \autoref{tab:res_table_all} shows the result of our experiment with three metrics.
% We can see the same result as in the main paper (Section 5.1) that our approach outperforms other models in detecting unknown outliers and also without a decrease in its performance on detecting known outliers.
% We see the same results when we use DAGMM and MemAE as baseline models where our approach outperform others in detecting both unknown and known outliers.

In pursuit of a comprehensive quantitative evaluation, we systematically assessed our methodology using various metrics, including 'F1-Score,' 'PR-AUC,' and 'ROC-AUC'. 
The results, summarized in \autoref{tab:res_table_all}, align with the findings presented in the main paper (Section 5.1). 
Our approach consistently exhibits superior performance in identifying unknown outliers, while concurrently sustaining its proficiency in detecting known outliers. 
This consistent trend persists when benchmarked against DAGMM and MemAE as baseline models, reaffirming our method's superior efficacy in both known and unknown outlier detection scenarios.

\begin{table*}[ht]
% \centering
% \begin{table}
% \begin{adjustbox}{\textwidth}
\resizebox{\textwidth}{!}{
\begin{tabular}{@{}llrrrrrr@{}}
\toprule
 &  & \multicolumn{2}{c}{\textbf{F1-Score}} & \multicolumn{2}{c}{\textbf{PR-AUC}} & \multicolumn{2}{c}{\textbf{ROC-AUC}} \\ \cmidrule(l){3-8} 
\textbf{Dataset} & \textbf{Model} & \textbf{\begin{tabular}[l]{@{}l@{}}Known\\ Outliers\end{tabular}} & \textbf{\begin{tabular}[l]{@{}l@{}}Unknown\\ Outliers\end{tabular}} & \textbf{\begin{tabular}[l]{@{}l@{}}Known\\ Outliers\end{tabular}} & \textbf{\begin{tabular}[l]{@{}l@{}}Unknown\\ Outliers\end{tabular}} & \textbf{\begin{tabular}[l]{@{}l@{}}Known\\ Outliers\end{tabular}} & \textbf{\begin{tabular}[l]{@{}l@{}}Unknown\\ Outliers\end{tabular}} \\ \midrule
\multirow{12}{*}{\begin{tabular}[c]{@{}l@{}}Credit \\ Default\end{tabular}} & RF & 0.946 $\pm$ 0.004 & 0.022 $\pm$ 0.004 & 0.973 $\pm$ 0.005 & 0.238 $\pm$ 0.005 & 0.994 $\pm$ 0.001 & 0.748 $\pm$ 0.001 \\
 & Baseline\_RF & 0.641 $\pm$ 0.025 & 0.114 $\pm$ 0.009 & 0.868 $\pm$ 0.008 & 0.431 $\pm$ 0.011 & 0.975 $\pm$ 0.005 & 0.818 $\pm$ 0.008 \\
 & Fed\_Avg+RF & 0.848 $\pm$ 0.013 & 0.389 $\pm$ 0.035 & 0.926 $\pm$ 0.012 & 0.688 $\pm$ 0.028 & 0.989 $\pm$ 0.003 & 0.920 $\pm$ 0.007 \\
 & Fed\_Prox+RF & 0.844 $\pm$ 0.019 & \textbf{0.406 $\pm$ 0.032} & 0.927 $\pm$ 0.016 & \textbf{0.694 $\pm$ 0.014} & 0.988 $\pm$ 0.003 & \textbf{0.923 $\pm$ 0.004} \\ \cmidrule(l){2-8} 
 & MLP & 0.939 $\pm$ 0.008 & 0.130 $\pm$ 0.012 & 0.953 $\pm$ 0.006 & 0.239 $\pm$ 0.013 & 0.990 $\pm$ 0.003 & \textbf{0.655 $\pm$ 0.014} \\
 & Baseline\_MLP & 0.920 $\pm$ 0.011 & 0.148 $\pm$ 0.016 & 0.941 $\pm$ 0.005 & 0.197 $\pm$ 0.020 & 0.984 $\pm$ 0.002 & 0.558 $\pm$ 0.020 \\
 & Fed\_Avg+MLP & 0.922 $\pm$ 0.013 & 0.360 $\pm$ 0.048 & 0.940 $\pm$ 0.014 & 0.369 $\pm$ 0.051 & 0.984 $\pm$ 0.005 & 0.598 $\pm$ 0.041 \\
 & Fed\_Prox+MLP & 0.915 $\pm$ 0.008 & \textbf{0.368 $\pm$ 0.024} & 0.933 $\pm$ 0.008 & \textbf{0.375 $\pm$ 0.015} & 0.982 $\pm$ 0.003 & 0.604 $\pm$ 0.019 \\ \cmidrule(l){2-8} 
 & DAGMM & 0.361 $\pm$ 0.010 & \textbf{0.117 $\pm$ 0.010} & 0.489 $\pm$ 0.018 & 0.372 $\pm$ 0.018 & 0.882 $\pm$ 0.012 & 0.778 $\pm$ 0.012 \\
 & DAGMM+Ours & 0.369 $\pm$ 0.013 & 0.107 $\pm$ 0.018 & 0.513 $\pm$ 0.025 & \textbf{0.373 $\pm$ 0.012} & 0.892 $\pm$ 0.009 & \textbf{0.778 $\pm$ 0.007} \\ \cmidrule(l){2-8} 
 & MemAE & 0.389 $\pm$ 0.016 & 0.054 $\pm$ 0.016 & 0.567 $\pm$ 0.012 & 0.369 $\pm$ 0.012 & 0.914 $\pm$ 0.003 & 0.834 $\pm$ 0.003 \\
 & MemAE+Ours & 0.411 $\pm$ 0.016 & \textbf{0.224 $\pm$ 0.013} & 0.589 $\pm$ 0.024 & \textbf{0.597 $\pm$ 0.011} & 0.927 $\pm$ 0.008 & \textbf{0.893 $\pm$ 0.002} \\ \midrule
\multirow{12}{*}{\begin{tabular}[c]{@{}l@{}}Adult \\ Data\end{tabular}} & RF & 0.790 $\pm$ 0.028 & 0.070 $\pm$ 0.027 & 0.802 $\pm$ 0.012 & 0.171 $\pm$ 0.010 & 0.928 $\pm$ 0.007 & 0.678 $\pm$ 0.005 \\
 & Baseline\_RF & 0.596 $\pm$ 0.017 & 0.119 $\pm$ 0.031 & 0.696 $\pm$ 0.031 & 0.295 $\pm$ 0.025 & 0.901 $\pm$ 0.005 & 0.809 $\pm$ 0.009 \\
 & Fed\_Avg+RF & 0.711 $\pm$ 0.037 & 0.309 $\pm$ 0.037 & 0.751 $\pm$ 0.022 & \textbf{0.483 $\pm$ 0.014} & 0.915 $\pm$ 0.005 & 0.841 $\pm$ 0.007 \\
 & Fed\_Prox+RF & 0.702 $\pm$ 0.039 & \textbf{0.339 $\pm$ 0.042} & 0.751 $\pm$ 0.028 & 0.482 $\pm$ 0.021 & 0.913 $\pm$ 0.004 & \textbf{0.842 $\pm$ 0.005} \\ \cmidrule(l){2-8} 
 & MLP & 0.776 $\pm$ 0.032 & 0.140 $\pm$ 0.019 & 0.763 $\pm$ 0.025 & 0.218 $\pm$ 0.021 & 0.893 $\pm$ 0.007 & 0.634 $\pm$ 0.039 \\
 & Baseline\_MLP & 0.754 $\pm$ 0.038 & 0.119 $\pm$ 0.005 & 0.745 $\pm$ 0.030 & 0.154 $\pm$ 0.013 & 0.888 $\pm$ 0.010 & 0.570 $\pm$ 0.031 \\
 & Fed\_Avg+MLP & 0.764 $\pm$ 0.030 & 0.364 $\pm$ 0.036 & 0.754 $\pm$ 0.020 & 0.375 $\pm$ 0.022 & 0.897 $\pm$ 0.003 & 0.668 $\pm$ 0.022 \\
 & Fed\_Prox+MLP & 0.758 $\pm$ 0.036 & \textbf{0.368 $\pm$ 0.037} & 0.752 $\pm$ 0.022 & \textbf{0.380 $\pm$ 0.037} & 0.901 $\pm$ 0.006 & \textbf{0.680 $\pm$ 0.034} \\ \cmidrule(l){2-8} 
 & DAGMM & 0.271 $\pm$ 0.063 & 0.110 $\pm$ 0.054 & 0.400 $\pm$ 0.057 & 0.234 $\pm$ 0.042 & 0.828 $\pm$ 0.020 & 0.740 $\pm$ 0.015 \\
 & DAGMM+Ours & 0.268 $\pm$ 0.054 & \textbf{0.110 $\pm$ 0.030} & 0.404 $\pm$ 0.052 & \textbf{0.238 $\pm$ 0.019} & 0.834 $\pm$ 0.007 & \textbf{0.740 $\pm$ 0.016} \\ \cmidrule(l){2-8} 
 & MemAE & 0.281 $\pm$ 0.036 & 0.132 $\pm$ 0.026 & 0.421 $\pm$ 0.015 & 0.295 $\pm$ 0.017 & 0.860 $\pm$ 0.004 & 0.796 $\pm$ 0.003 \\
 & MemAE+Ours & 0.594 $\pm$ 0.072 & \textbf{0.279 $\pm$ 0.023} & 0.692 $\pm$ 0.052 & \textbf{0.498 $\pm$ 0.034} & 0.901 $\pm$ 0.008 & \textbf{0.860 $\pm$ 0.001} \\ \midrule
\multirow{12}{*}{MNIST} & RF & 0.838 $\pm$ 0.039 & 0.323 $\pm$ 0.032 & 0.962 $\pm$ 0.008 & \textbf{0.743 $\pm$ 0.005} & 0.994 $\pm$ 0.004 & 0.930 $\pm$ 0.003 \\
 & Baseline\_RF & 0.502 $\pm$ 0.090 & 0.144 $\pm$ 0.023 & 0.817 $\pm$ 0.070 & 0.534 $\pm$ 0.083 & 0.966 $\pm$ 0.047 & 0.919 $\pm$ 0.043 \\
 & Fed\_Avg+RF & 0.660 $\pm$ 0.067 & 0.224 $\pm$ 0.108 & 0.890 $\pm$ 0.010 & 0.677 $\pm$ 0.066 & 0.994 $\pm$ 0.004 & 0.975 $\pm$ 0.009 \\
 & Fed\_Prox+RF & 0.703 $\pm$ 0.057 & \textbf{0.326 $\pm$ 0.083} & 0.898 $\pm$ 0.019 & 0.718 $\pm$ 0.031 & 0.994 $\pm$ 0.005 & \textbf{0.976 $\pm$ 0.004} \\ \cmidrule(l){2-8} 
 & MLP & 0.948 $\pm$ 0.009 & 0.424 $\pm$ 0.026 & 0.981 $\pm$ 0.006 & \textbf{0.656 $\pm$ 0.014} & 0.999 $\pm$ 0.001 & \textbf{0.944 $\pm$ 0.008} \\
 & Baseline\_MLP & 0.782 $\pm$ 0.179 & 0.381 $\pm$ 0.104 & 0.873 $\pm$ 0.116 & 0.516 $\pm$ 0.116 & 0.974 $\pm$ 0.032 & 0.890 $\pm$ 0.042 \\
 & Fed\_Avg+MLP & 0.822 $\pm$ 0.049 & 0.399 $\pm$ 0.104 & 0.910 $\pm$ 0.019 & 0.562 $\pm$ 0.088 & 0.992 $\pm$ 0.004 & 0.932 $\pm$ 0.028 \\
 & Fed\_Prox+MLP & 0.879 $\pm$ 0.021 & \textbf{0.482 $\pm$ 0.059} & 0.929 $\pm$ 0.019 & 0.614 $\pm$ 0.066 & 0.996 $\pm$ 0.004 & 0.933 $\pm$ 0.025 \\ \cmidrule(l){2-8} 
 & DAGMM & 0.377 $\pm$ 0.102 & 0.091 $\pm$0.100 & 0.669 $\pm$ 0.039 & 0.354 $\pm$ 0.035 & 0.957 $\pm$ 0.002 & 0.868 $\pm$ 0.002 \\
 & DAGMM+Ours & 0.480 $\pm$ 0.094 & \textbf{0.200 $\pm$ 0.101} & 0.673 $\pm$ 0.044 & \textbf{0.415 $\pm$ 0.113} & 0.955 $\pm$ 0.009 & \textbf{0.887 $\pm$ 0.029} \\ \cmidrule(l){2-8} 
 & MemAE & 0.846 $\pm$ 0.013 & \textbf{0.211 $\pm$ 0.011} & 0.955 $\pm$ 0.007 & 0.574 $\pm$ 0.007 & 0.997 $\pm$ 0.003 & 0.930 $\pm$ 0.001 \\
 & MemAE+Ours & 0.697 $\pm$ 0.067 & 0.178 $\pm$ 0.056 & 0.924 $\pm$ 0.006 & \textbf{0.699 $\pm$ 0.040} & 0.994 $\pm$ 0.002 & \textbf{0.982 $\pm$ 0.003} \\ \bottomrule 
\end{tabular}
}
% \end{adjustbox}
\caption{Different evaluation metrics score for baseline and FL-OD Models. Our approach outperform others in detecting both unknown and known outliers. Every score reflects the mean and standard deviation from five experiments.}
\label{tab:res_table_all}
% \end{table}
\end{table*}

\newpage
\section{Quantitative Evaluation Clientwise Result}
% Below figures show the clientwise result comparison between the baseline and our approach for known and unknown outlier detection using different evaluation metrics.
% We used two types of OD algorithms namely `Random Forest (RF)' and `Multi Layer Perceptron (MLP)'
% In most of the cases our federated learning based approach i.e, Fed\_Avg+RF, Fed\_Prox+RF, Fed\_Avg+MLP, Fed\_Prox+MLP outperformed other approaches in detecting unknown outliers.

\begin{figure}[!htb]
% \begin{figure*}[ht]
\centering
\includegraphics[width=0.6\linewidth]{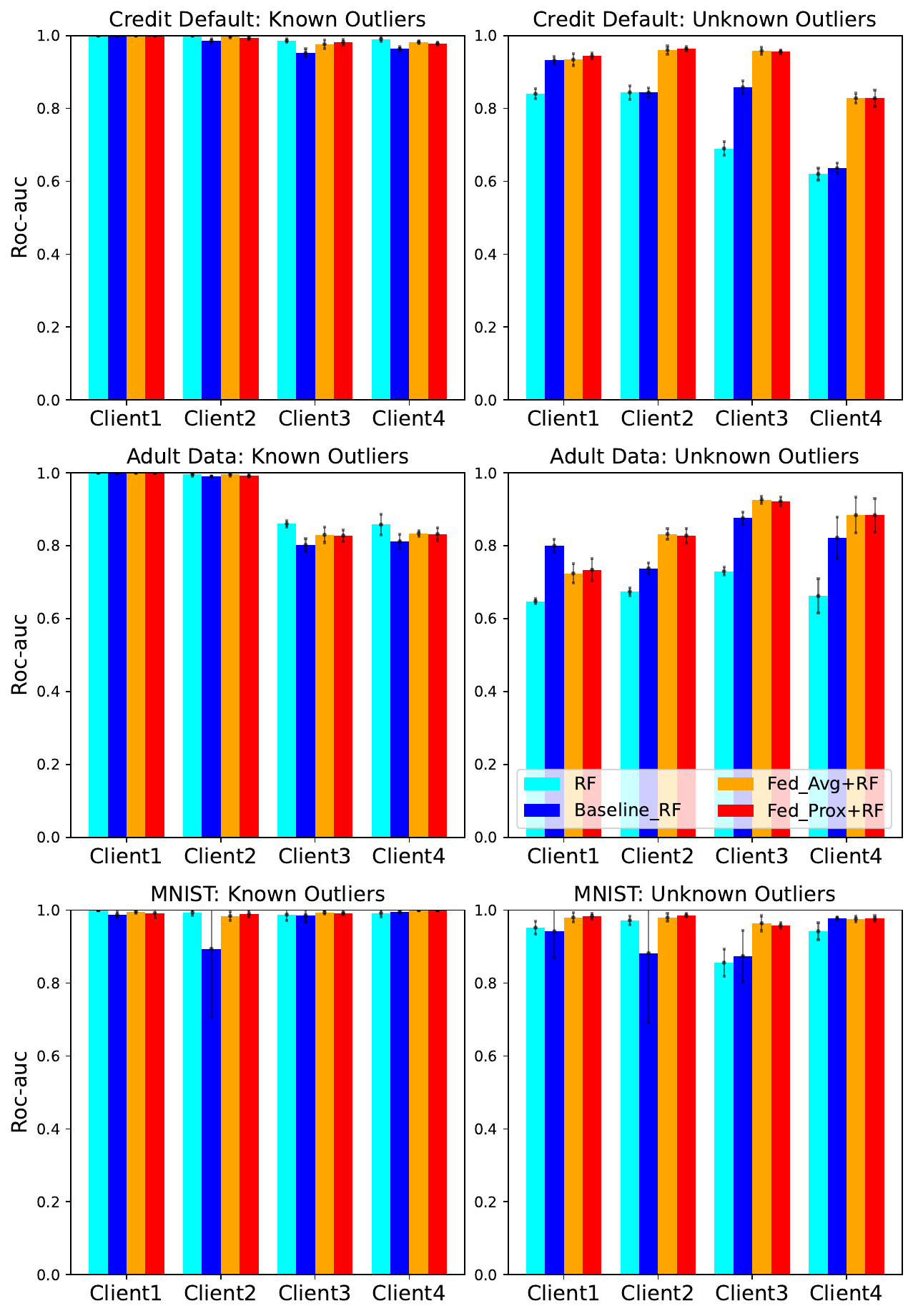}
\caption{Results of OD models clientwise. FL-OD models (Fed\_Avg+RF, Fed\_Prox+RF) outperform the baseline (Baseline\_RF) in detecting unknown outliers. Every score reflects the mean and standard deviation from five experiments.}
\label{fig:Fig1}
\end{figure}

\begin{figure}[!htb]%
    \centering
    \subfloat[\centering F1-Score]{{\includegraphics[width=0.55\linewidth]{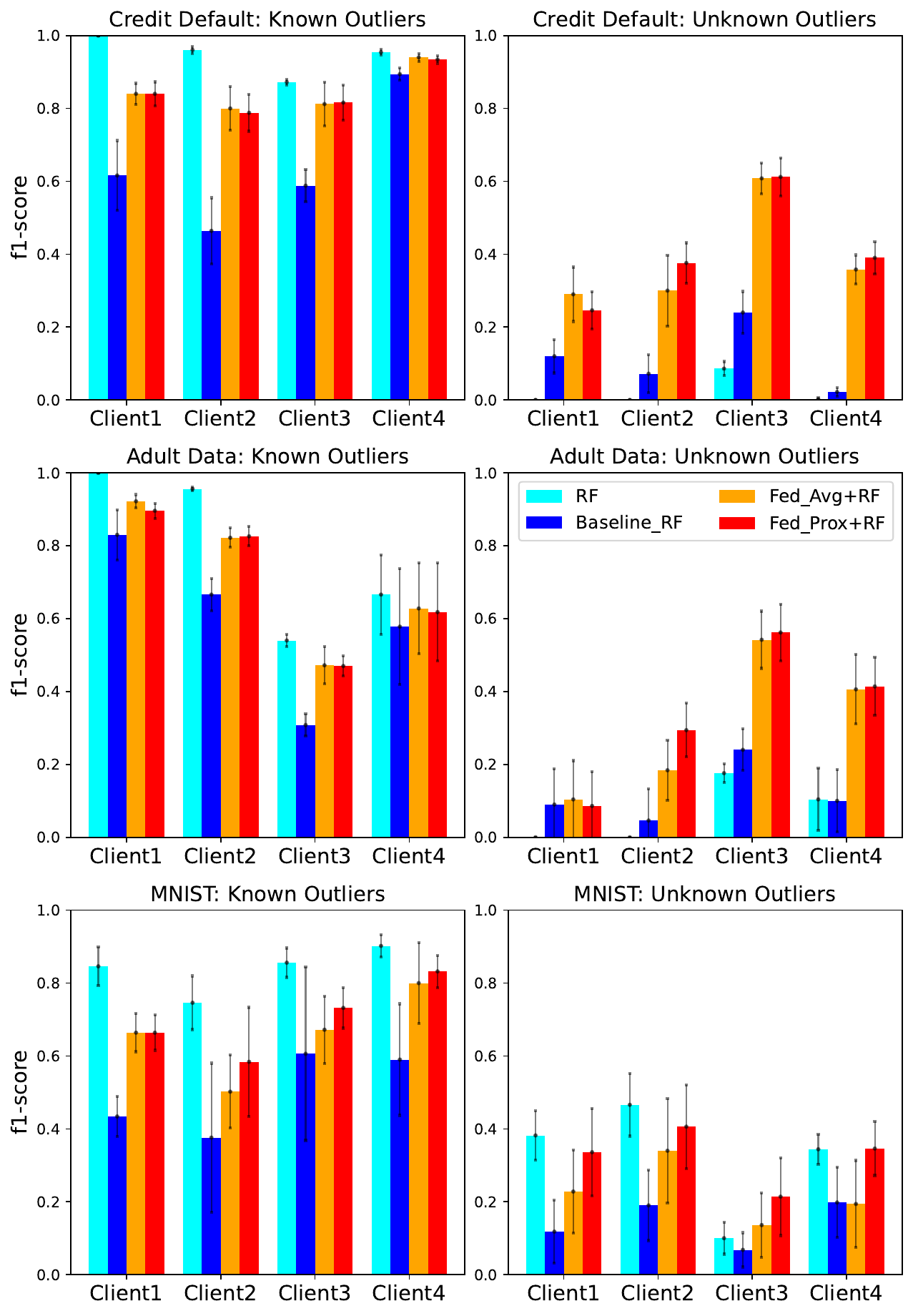} }}%
    % \qquad
    \subfloat[\centering PR-AUC]{{\includegraphics[width=0.55\linewidth]{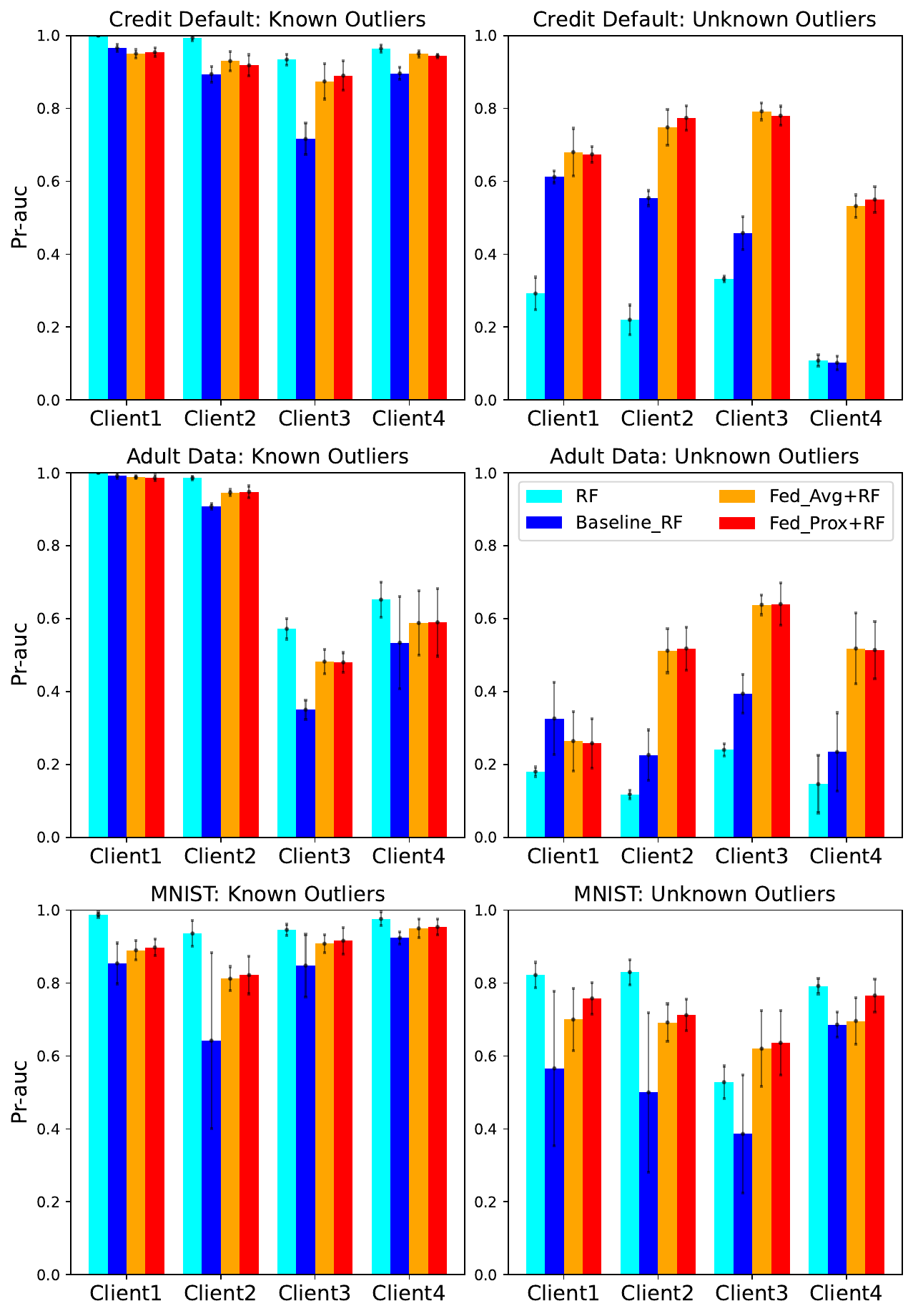} }}%
    \caption{Results of OD models clientwise. FL-OD models (Fed\_Avg+RF, Fed\_Prox+RF) outperform the baseline (Baseline\_RF) in detecting unknown outliers. Every score reflects the mean and standard deviation from five experiments.}%
    \label{fig:Fig2}%
\end{figure}

\begin{figure}[!htb]%
    \centering
    \subfloat[\centering Average Precision Score]{{\includegraphics[width=0.55\linewidth]{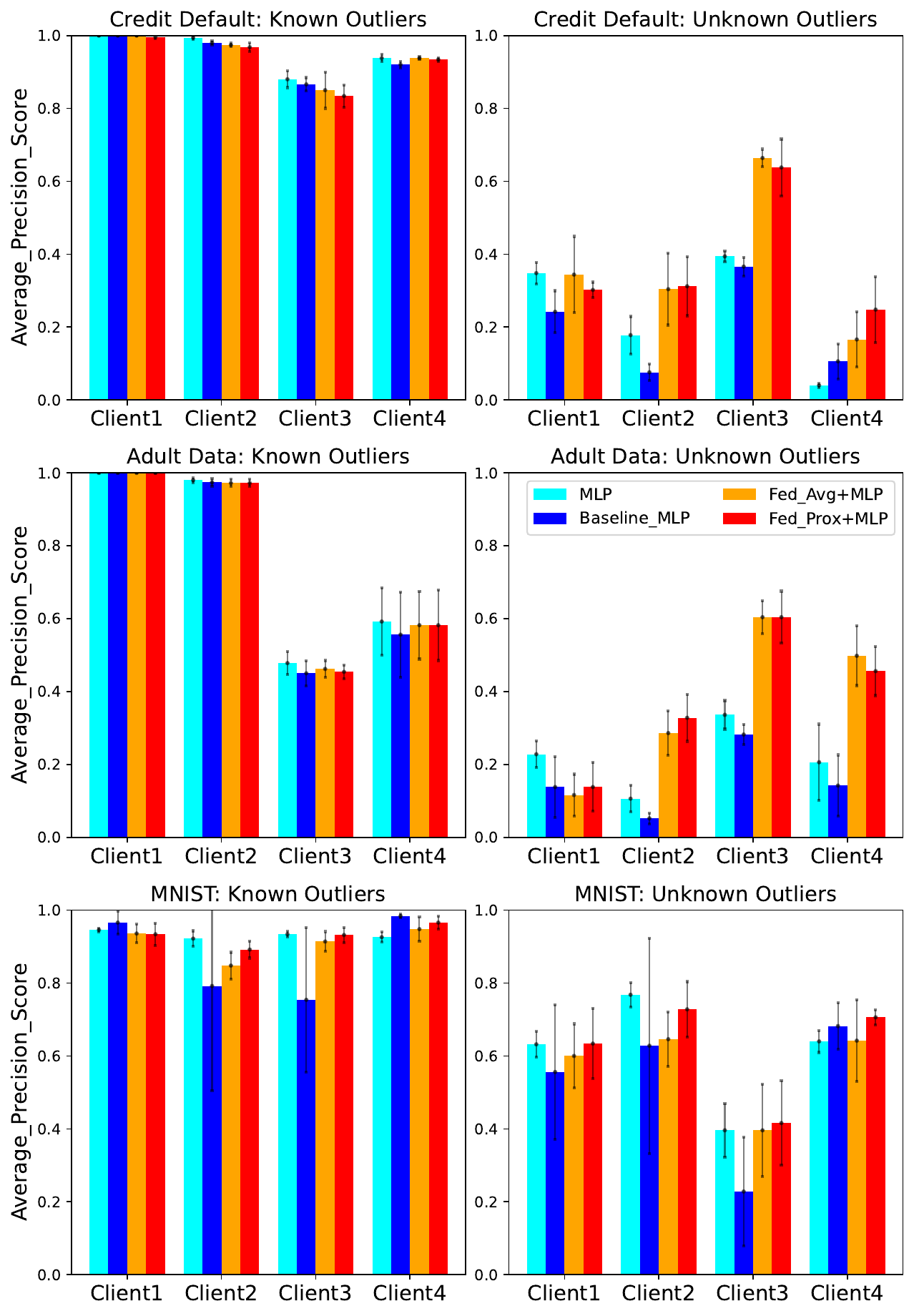} }}%
    % \qquad
    \subfloat[\centering F1-Score]{{\includegraphics[width=0.55\linewidth]{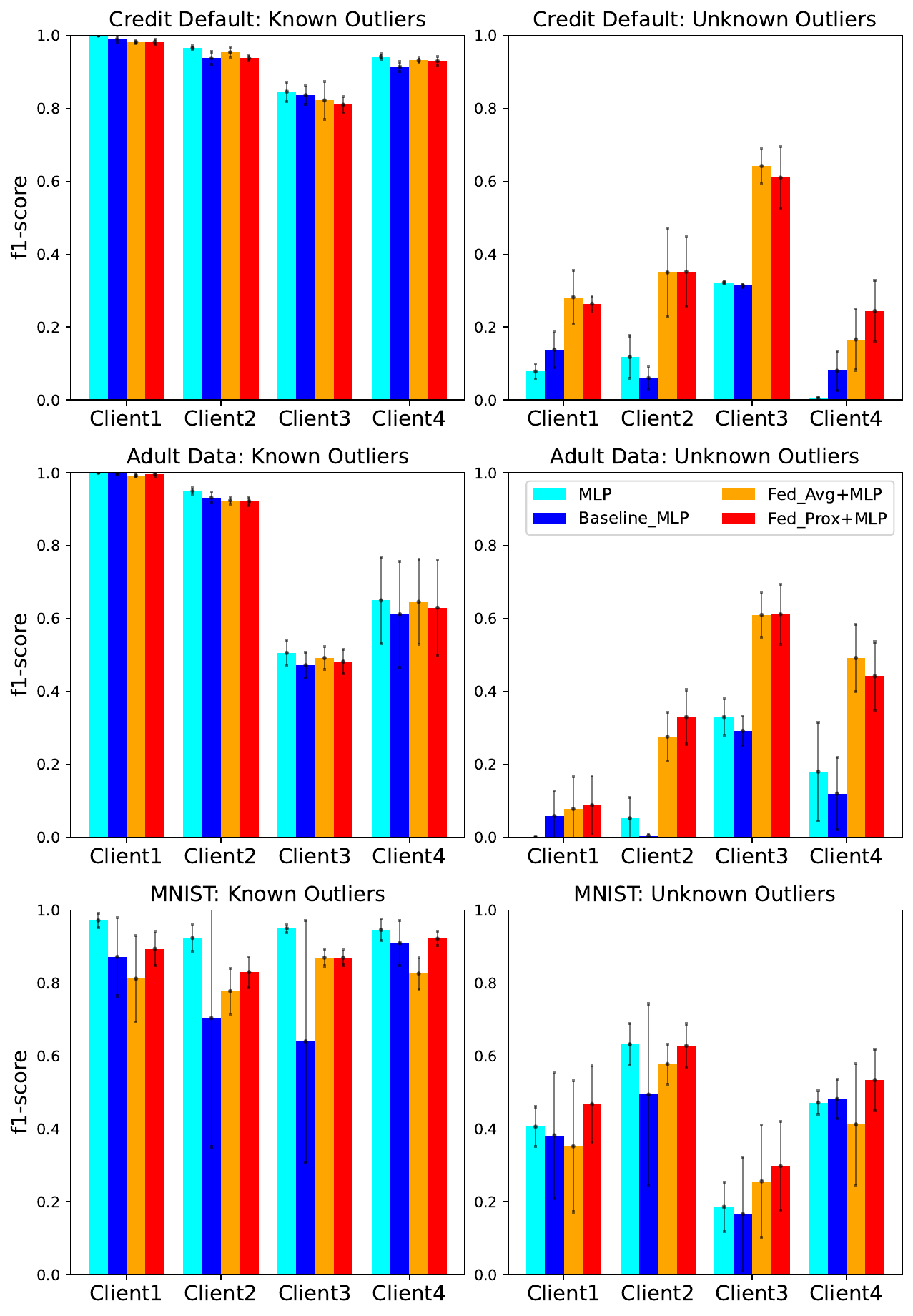} }}%
    \caption{Results of OD models clientwise. FL-OD models (Fed\_Avg+MLP, Fed\_Prox+MLP) outperform the baseline (Baseline\_MLP) in detecting unknown outliers. Every score reflects the mean and standard deviation from five experiments.}%
    \label{fig:Fig3}%
\end{figure}

\begin{figure}[!htb]%
    \centering
    \subfloat[\centering PR-AUC]{{\includegraphics[width=0.55\linewidth]{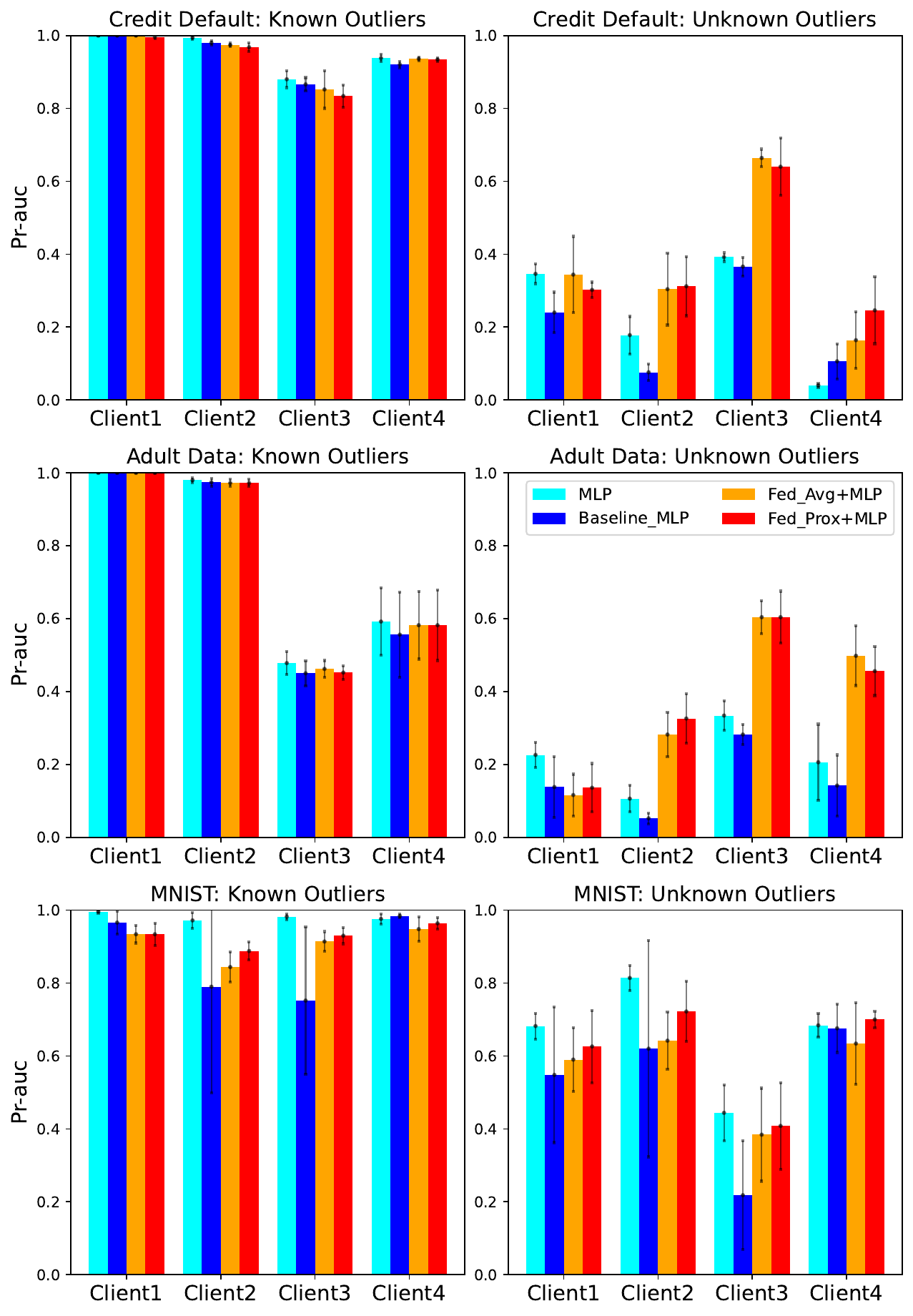} }}%
    % \qquad
    \subfloat[\centering PR-AUC]{{\includegraphics[width=0.55\linewidth]{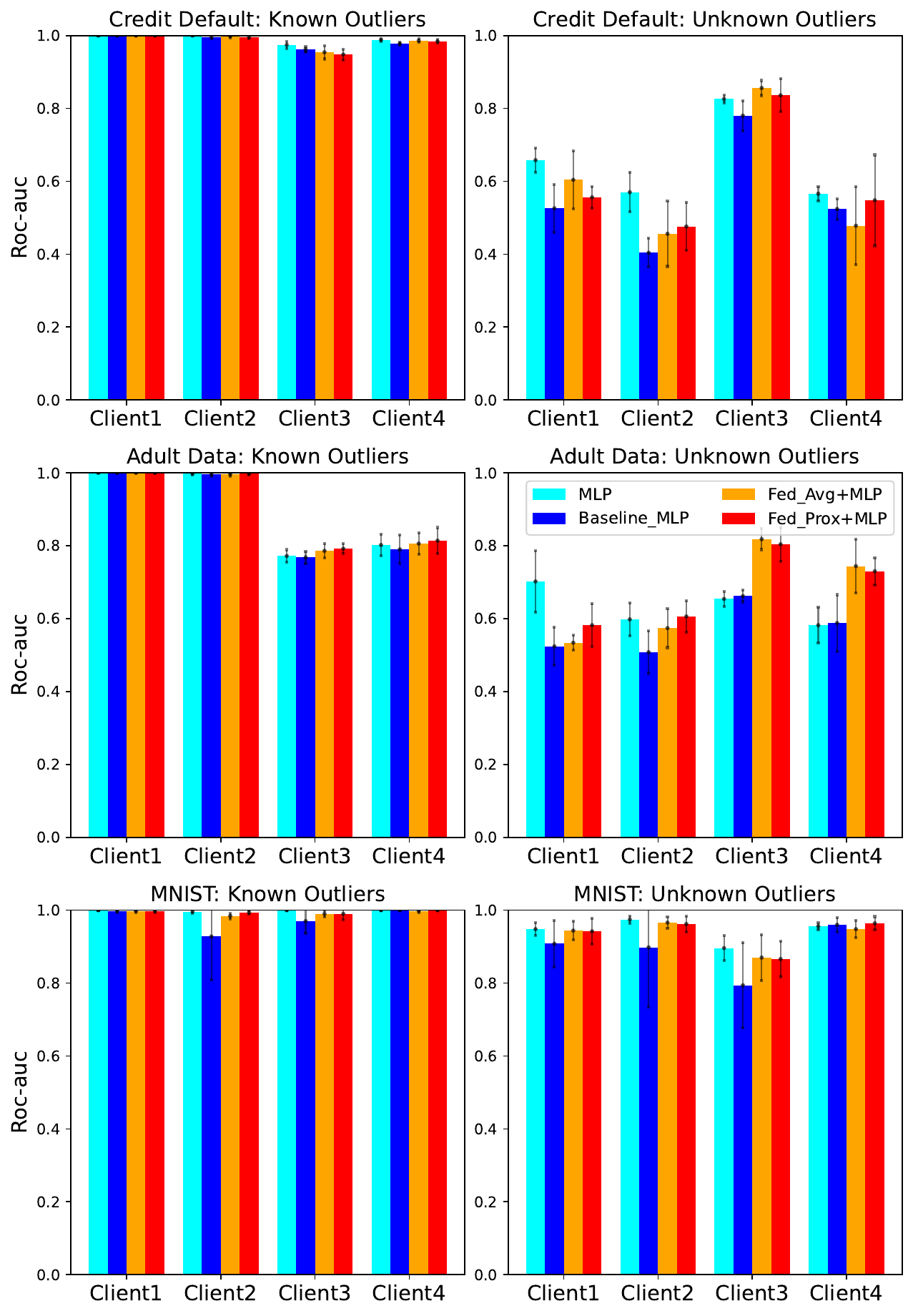}}}%
    \caption{Results of OD models clientwise. FL-OD models (Fed\_Avg+MLP, Fed\_Prox+MLP) outperform the baseline (Baseline\_MLP) in detecting unknown outliers. Every score reflects the mean and standard deviation from five experiments.}%
    \label{fig:Fig4}%
\end{figure}

The figures below (\autoref{fig:Fig1}, \autoref{fig:Fig2}, \autoref{fig:Fig3}, \autoref{fig:Fig4}) depict a detailed clientwise comparison of results between the baseline and our proposed methodology, specifically focusing on both known and unknown outlier detection and utilizing a range of evaluation metrics. 
Two distinct outlier detection (OD) algorithms, 'Random Forest (RF)' and 'Multi Layer Perceptron (MLP),' were employed for this comprehensive analysis. 
Remarkably, across various scenarios, our federated learning-based approach—encompassing Fed\_Avg+RF, Fed\_Prox+RF, Fed\_Avg+MLP, and Fed\_Prox+MLP—consistently demonstrated superior performance in detecting unknown outliers when compared to alternative methods.

\subsection{OD\_Model: Random Forest (RF)}

% \end{figure*}

\clearpage

\newpage
\subsection{OD\_Model: Multi Layer Perceptron (MLP)}

\clearpage

\newpage
\section{Model Description}

\subsection[]{Deep Auto Encoding Gaussian Mixture Model (DAGMM) \cite{DAGMM}}:

\begin{table}[H]
\centering
% \begin{table}
% \begin{adjustbox}{\textwidth}
% \resizebox{\textwidth}{!}{
\begin{tabular}{@{}lrrrrr@{}}
\toprule
 & \multicolumn{5}{c}{DAGMM} \\ \midrule
Dataset & Batch & Epoch & n\_GMM & Learning Rate & Weight Decay \\ \midrule
Credit Default & 1024 & 500 & 4 & 0.0001 & 0.0001 \\
Adult Data & 1024 & 500 & 4 & 0.0001 & 0.0001 \\
MNIST & 128 & 1000 & 2 & 0.0001 & 0.0001 \\ \bottomrule
\end{tabular}
% }
% \end{adjustbox}
\caption{DAGMM Hyperparameters}
\label{tab:dagmm_hyperparameters}
% \end{table}
\end{table}

\subsection[]{Memory-augmented Deep Autoencoder (MemAE) \cite{memae}}:

\begin{table}[H]
\centering
% \begin{table}
% \begin{adjustbox}{\textwidth}
% \resizebox{\textwidth}{!}{
\begin{tabular}{@{}lrrrrr@{}}
\toprule
 & \multicolumn{5}{c}{MemAE} \\ \midrule
Dataset & \multicolumn{1}{l}{Batch} & \multicolumn{1}{l}{Epoch} & \multicolumn{1}{l}{Memory Dimension} & \multicolumn{1}{l}{Learning Rate} & \multicolumn{1}{l}{Weight Decay} \\ \midrule
Credit Default & 1024 & 500 & 100 & 0.0001 & 0.0001 \\
Adult Data & 1024 & 500 & 100 & 0.0001 & 0.0001 \\
MNIST & 1024 & 1000 & 50 & 0.0001 & 0.0001 \\ \bottomrule
\end{tabular}
% }
% \end{adjustbox}
\caption{MemAE Hyperparameters}
\label{tab:dagmm_hyperparameters}
% \end{table}
\end{table}

\end{document}